# Text-Independent Writer Identification using Convolutional Neural Networks

**Hung T. Nguyen**
Department of Computer and Information Science
Tokyo University of Agriculture and Technology (TUAT)
Tokyo, Japan 184-8588
`ntuanhung@gmail.com`

**Cuong T. Nguyen**
Dept. of Comp. and Info. Science
TUAT
Tokyo, Japan 184-8588
`ntcuong2103@gmail.com`

**Takeya Ino**
Dept. of Comp. and Info. Science
TUAT
Tokyo, Japan 184-8588
`ntuanhung@gmail.com`

**Bipin Indurkhya**
Institute of Philosophy
Jagiellonian University
Cracow, Poland
`bipin.indurkhya@uj.edu.pl`

**Masaki Nakagawa**
Dept. of Comp. and Info. Science
Tokyo, Japan 184-8588
`nakagawa@cc.tuat.ac.jp`

## Abstract

This paper proposes an end-to-end deep-learning based method for text-independent writer identification. In this approach, convolutional neural networks (CNNs) are trained initially to extract the local features which represent characteristics of individual handwriting in the whole and sub-regions of character images. We make randomly sampled tuples of images from the training set to train CNNs and aggregate the extracted local features of images from the tuples to form the global features. By alternating the images to make the tuples, we create a large number of training patterns required by text-independent writer identification as well as the training process of CNNs. Experiments on the JEITA-HP database of offline handwritten Japanese character patterns show the effectiveness of this approach to overcome the difficulties of gathering handwritten character patterns of the same categories as the specimens of the writer. When we can use 200 characters, the method realizes the accuracy of 99.97% to classify 100 writers. Even if we can only use 50 characters, the method achieves the accuracy of 92.8%, which shows the ability to retain the accuracy despite the more number of writers or the less number of characters for training. Moreover, we made experiments on the Firemaker database and the IAM database of offline handwritten English text. When we use one page per writer to train, the method exceeds the accuracy of 91.5% to classify 900 writers. This result, as well as results for other conditions, show better performance than the previously published best result using handcrafted features and clustering algorithms, which shows the effectiveness of the method also for handwritten English text.

# 1 Introduction

Writer identification has been studied for many years because it has practical applications for forgery detection and forensic science. Since the mid-1960s, computational approaches have been developed for writer identification. A review of the state-of-the-art methods for writer verification and identification from the 1960s to the 1980s can be found in Plamondon and Lorette (1989). The researchers in the early days of this research focused on using offline signatures to identify the writer [2], [3] because the signatures were considered as individual signs from the past. Yoshimura et al. proposed a text-dependent writer identification method using handwritten text [4], [5]. Due to the difficulty of collecting multiple signature patterns and the availability of large databases of handwritten text from the early 2000s, there have been many studies on writer identification employing handwritten text. Srihari et al. claimed that the individuality of handwritten text is appropriate for the writer identification task and machine learning algorithms can be trained for this task with a large number of writers [6].

There are two main approaches for writer identification: text-dependent and text-independent approaches. The text-dependent approach demands the same text to be written while the text-independent approach does not require any particular text. The writer identification research so far has focused on deciding whether the given handwritten characters are written by the person whose handwritten sample patterns of the same categories are available. The problem with this approach is that we cannot always find the sample patterns of the same categories as the target patterns. The text-independent approach does not require collecting the same category patterns, which is useful in situations like forensic science. However, the text-independent approach is more complicated than the text-dependent approach because it must extract writer-specific features regardless of signatures, specified letters, or symbols to be compared. In the research presented here, we focus on solving text-independent writer identification task.

Recently, Sreeraj and Idicula reviewed both the text-dependent and text-independent writer identification methods from the late 1990s to 2010 [7]. These methods employed textual features [8], edge-based features [9] and allograph features [10]. In addition, they reviewed many approaches for identifying the writer based on the extracted features: for instance, cosine similarity [11], k-nearest neighbor [12] and clustering methods [13]. Among these methods, the approach by Bulacu and Schomaker (2007) provides a the theoretical foundation for writer identification based on handwritten text, where each handwriting pattern is described by a two-level psychomotor process. In this approach, the handwritten patterns are produced from a complicated sequence of thinking and hand movements, which are characteristic of each writer depending on his/her cognitive system, nervous system, muscle system, gender, age, schooling, and so on. Therefore, each writer has his/her allographs whenever he/she writes a letter which suggests that it is possible to identify a person based on her or his writing behavior. Bulacu and Schomaker (2007) conducted experiments on the handwriting databases with 900 writers using their handcrafted features and clustering methods. The best accuracy results achieved were 80% when using a single handcrafted feature and 87% when using combined features.

There are two kinds of handwritten text patterns: online patterns (time series of pen trajectories to write text) and offline patterns (handwritten text images). Consequently, there are online methods and offline methods of writer identification. Moreover, two kinds of features are usually employed for writer identification: local features and global features. For offline writer identification, local features are extracted from sub-regions of an image based on local descriptors such as scale-invariant features transform [14], [15], local binary pattern [16], and contour features [17]. Global features are extracted from an input image at the document level and paragraph level using ink width [18], structural features [19], and texture features [11], [20]. Moreover, there are several studies on combining local and global features [6], [10], [21]. The approach presented in this paper deals with offline writer identification using the trained features from CNNs.

The rest of this paper is organized as follows: Section 2 presents a brief survey of recent research on text-independent writer identification using CNNs. Section 3 presents the details of our method, which incorporates local feature extraction, feature aggregation to form global features and sampling mechanism. Section 4 presents experimental results demonstrating the efficacy of our proposed method on Japanese and English handwriting databases. In Section 4, our method is compared with the best existing approach that uses handcrafted features and clustering methods on English databases. Finally, Section 5 presents our conclusions and suggestions for future research.

## 2      Related works

As a text-independent approach requires writer-specific features, hand-crafted features are difficult to define. Therefore, it is highly desirable to automatically detect effective writer-specific features from handwritten patterns. During the last decade, deep learning techniques have been successfully applied to many recognition tasks [22]–[24] because they are efficient in automatic learning of features.

For the writer identification task, Fiel and Sablatnig (2015) employed CNNs as local feature extractors. In their approach, the last fully connected layer was eliminated because the layer just above it extracted adequate features to identify the writer. Then, the mean vector for the input image was computed using all its local feature vectors, which was used for identifying the writer based on Chi-Square distance. This method requires a preprocessing step for image binarization and normalization, so its performance depends on the database and the preprocessing method.

Christlein et al. (2015) presented another strategy: after extracting local features by CNNs, they are used to form global features based on Gaussian Mixture Models (GMM) super-vector encoding. This combination method performs better than that of Fiel and Sablatnig (2015). Moreover, CNNs achieve as better performance than the traditional local descriptors such as SIFT.

Yang et al. (2016) proposed an end-to-end online text-independent writer identification system, which generates many artificial patterns from a single original online character patterns by dropping one or more strokes. The artificial patterns formed by the same original pattern are fed into CNNs to get a probability distribution of each character. Then, the average distributions of all the character patterns in the document are computed and used for identifying the writer. Even though this system achieves a high accuracy, it requires many character patterns in documents to make the average reliable. Moreover, its DropSegment method is difficult to apply for offline handwritten text patterns.

The systems of Christlein et al. (2015) and Fiel and Sablatnig (2015) have two separate training steps: feature extraction and encoding where CNNs are pre-trained and employed for local features extraction. The pre-trained CNNs are fixed during the second step while training the encoder, which reduces the performance of the whole system because the CNNs are not updated. Therefore, we propose to use end-to-end networks for writer identification so that both the feature extractor and the classifier are trained together.

## 3      Proposed method

We propose here an end-to-end method based on deep learning, which extracts features using CNNs and combines the extracted features from handwritten text images of multiple characters as shown in Fig. 1. As our method, which forms tuples of multiple randomly sampled characters as input during training process, uses a new way of organizing training samples, CNNs are able to extract text-independent writer-specific features. Also, we apply some different feature aggregation methods for forming global features from the extracted local features. The aggregated features are fed to a fully connected layer consisting of $N_{fc}$ units which equals the number of writers, to make a prediction. The whole network is trained by the stochastic gradient descent algorithm, which helps to discover not only the local features but also the global features that are hard to define by handcrafted features.

Our approach incorporates both local and global features, which are text independent, i.e., character category invariant. We consider two levels of local features: the sub-region level and the character level. The sub-region level, extracted from sub-regions of a character image, captures writer-specific features in writing strokes, which are directly related to the psychomotor process to identify the writer. The character level, extracted from a character image, captures features related to writing styles such as character balancing, character layouts or stroke combinations.

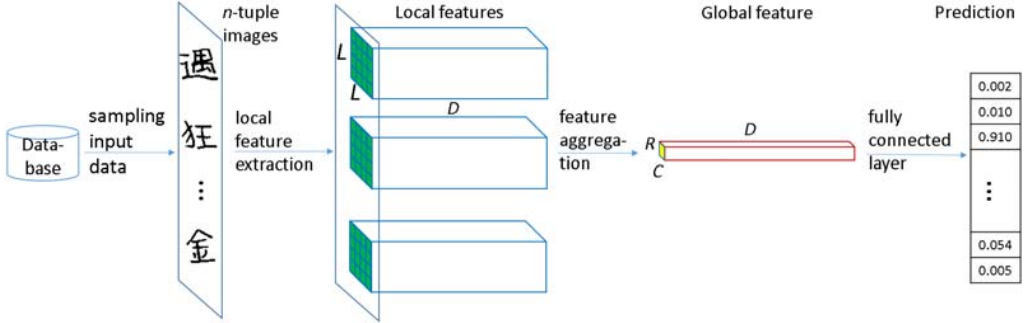

Figure 1: Overview of our proposed method.

### 3.1 Local feature at the sub-region level

We use the model illustrated in Fig. 2 (a) for extracting features from sub-regions in a character image. Each convolution layer uses a kernel size of 5x5, stride step of 2 and padding size of 2. Each max-pooling layer uses a kernel size of 2x2 and stride step of 2. We use 4 blocks of a convolutional layer followed by a max-pooling layer to produce the 4x4 feature maps from a 64x64 input image. Each column in these feature maps can be considered as the features from a sub-region of an input character image as shown in Fig. 2 (b). Specifically, each column of the 4x4 feature maps represents the features extracted from a 16x16 sub-region of the entire character image. These local features (at the sub-region level) of each character image are extracted as a 4x4x1024-dimensional vector.

### 3.2 Local features at the character level

For extracting features at the character level, we use 3 blocks of a convolutional layer followed by a max-pooling layer to produce 8x8 feature maps from a 64x64 input image as illustrated in Fig. 3. These feature maps are then fed to a fully connected layer to extract features of the entire image of the input character. Consequently, these local feature (at the character level) of each character are extracted and represented by a 1x1x1024-dimensional vector. Thus, all local features could be represented as an $L \times L \times D$ vector as shown in Fig. 1.

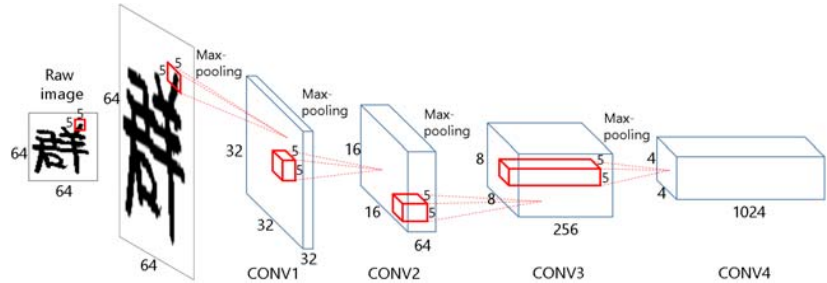

**(a)** The model architecture consists of 4 blocks of a convolution layer followed by a max-pooling layer with the numbers of filters being 32, 64, 256, and 1024, respectively. The red block in each convolution layer indicates the convolution kernel.

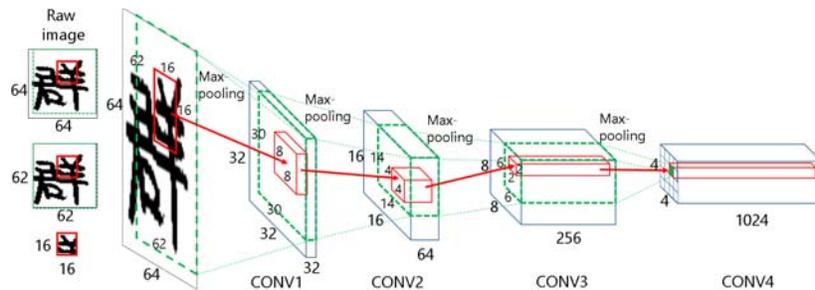

**(b)** The red block of CONV1 is extracted features from the 16x16 sub-region of an input image. Each red block of other convolution layers indicates the features from the red block of its previous convolution layer. Each green block in the convolution layer shows the bounding box from which the sub-region image features could be extracted.

Figure 2: CNN model for extracting sub-regions level based local features.

### 3.3 Global features from multiple characters

Local features from a single character image may not contain enough information for identifying the writer. Therefore, we consider extracting global features for writer identification by aggregating local features. Instead of using the whole document or paragraph, we extract global features from multiple character images. Global features such as structural features, slant, and so on may vary from one character image to another, but we propose that they can be extracted reliably from multiple character images. By design two following aggregation methods, we tend to extract writer-specific features from local features.

Fig. 4 shows the general process for obtaining global features from *n*-tuple images which are fed into CNNs to extract local features. Thus, the local feature vector of each input image is of the size $L \times L \times D$. Next, an aggregation method is applied in order to get the global feature vector which is of the size $R \times C \times D$ where $R$ is the number of rows, $C$ is the number of columns and $D$ is the depth.

We consider two basic methods for gathering the local features (features extracted from the sub-region level and the character level) to form global features (features from multiple characters) as shown in Fig. 4. One method is "Average Aggregation" (AA), which works by computing the average of all the values for each dimension of the feature vectors, as shown in Eq. (1). AA produces more robust global features compared with the local features.

$$\text{global\_feature}[d]_{1 \leq d \leq D} = \frac{1}{L^2} \left( \sum_{i}^{L} \sum_{j}^{L} \text{local\_feature}[i, j, d] \right) \quad (1)$$

The other method is "Max Aggregation" (MA), which works by selecting the maximum value along each dimension, as shown in Eq. (2). MA selects the most relevant local feature in each feature dimension for writer identification. Both the AA and MA global features vectors are of the size 1x1x1024.

$$\text{global\_feature}[d]_{1 \leq d \leq D} = \max_{1 \leq i, j \leq L} \{\text{local\_feature}[i, j, d]\} \quad (2)$$

For the AA method, the common features shared among them are made more robust by averaging the local features. On the other hand, the MA method can discover relevant features that may not appear in all the local features in some dimension. The difference between the above two methods suggests that combining them may result in a better aggregation. We also consider incorporating the method of "K-Max Aggregation" (KMA), which retains the top $K$ maximum values and averages them to yield a global feature as shown in Eq. (3).

Thus, the KMA global features vector is of the size $K \times 1 \times 1024$.

$$\text{global\_feature}[d]_{1 \leq d \leq D} = \text{top}-k-\max_{1 \leq i, j \leq L} \{\text{local\_feature}[i, j, d]\} \quad (3)$$

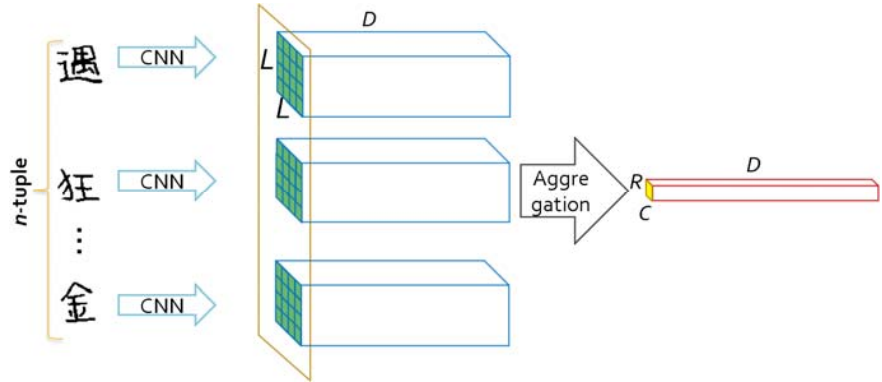

Figure 4: Global features from multiple characters.

### 3.4 Random sampling of training patterns

We assume at least $Ns$ character images are available. Each training iteration, except the initial iteration, consists of a random permutation of the $Ns$ images and separating the permuted $Ns$ images into $m$ sets of $n$-tuples of multiple character images. The training iteration is looped $p$ times for each writer as shown in Fig. 5, and we consider this set of iterations as an epoch. The training epochs are repeated until there is no improvement in the validation accuracy. By reordering images of n-tuples in each iteration, and in each epoch, the layers of networks are trained to extract writer-specific features without text dependency.

## 4 Experiments

### 4.1 Databases

#### 4.1.1 JEITA-HP database

We employ the JEITA-HP handwritten Kanji character pattern database (hereafter, JEITA-HP), which is prepared by Hewlett-Packard Japan and distributed by Japan Electronics and Information Technology Industries Association (JEITA) [27]. The JEITA-HP database consists of two datasets: Dataset A and Dataset B which store handwritten character patterns from 480 writers and 100 writers, respectively. In JEITA-HP, each writer provided 3,306 patterns of 3,214 categories (2,965 Kanji, 82 Hiragana, 10 numerals, 157 other characters: English alphabet, Katakana and symbols) in which each Kanji character was written once. In our experiments, we only use Kanji character patterns from the writers in Dataset A. Firstly, we split 2,965 Kanji characters into 3 subsets 2,000 characters for the training set, 400 characters for the validation set and the other characters for the testing set. We randomly select writers and training samples from the training set based on the number of writers and the number of characters for training which depends on each experiment.

#### 4.1.2 Firemaker and IAM databases

In order to show our method for English and to compare our method with the handcrafted features in [10], we employ the Firemaker and IAM databases of handwritten text. For the Firemaker database provided by 250 writers, there are four subsets which are collected by different requirements [28]. The first subset contains the text-copying pages using normal handwriting which is used as the training and validation sets. The second subset contains the text-copying pages using only uppercase characters, which is unusual handwritten text and thus not used in our experiments. The third subset contains forged text, where the writers were asked to write in different style. Hence, the third subset is not used in our experiments. The fourth subset contains the free handwritten text to describe the content of a given cartoon which is used as the testing set.

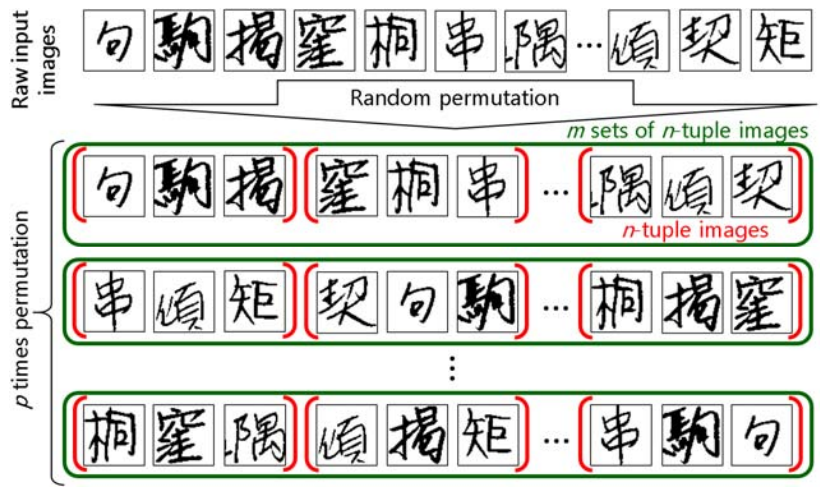

Figure 5: Randomly sampled training images.

For the IAM database provided by 650 writers, each writer provided from 1 page (350 writers) to the largest 59 pages (1 writer) [29]. Due to the imbalance of patterns from writers, we did the following process to balance the patterns of each writer. For the writers who only provided 1 page, their pages were divided roughly into half: the first half for training/validation and the second half for testing. For the writers who provided more than one pages, only their first two pages were selected with the first page for training/validation and the second page for testing.

For preprocessing pages from the Firemaker and IAM databases, we employ the Otsu binarization method. The handwritten text in these two databases is cursively written which is difficult to segment and separate each character. Fortunately, our proposed method is aimed to extract the writer-specific features based on the sub-regions of characters which implies that it does not require isolated characters as input. Thus, we do not need to segment the handwritten text from the Firemaker and IAM databases.

Instead, we extract $N_{sub\_img}$ sub-images of $k_{sub\_img}$ x $k_{sub\_img}$ pixels from the binarized page based on the handwritten text probability as shown in Fig. 6. The handwritten text probability is obtained by applying the average filter with a size of 32x32 pixels and then normalizing them to the appropriate range for representing the probability. In our experiment, we employ 500 for $N_{sub\_img}$ and 64 for $k_{sub\_img}$ since they could cover all the handwritten text of the page. Similar to the experiments on JEITA-HP, we randomly select writers and training samples from the training set based on the number of writers and the number of sub-images for training which depends on each experiment.

## 4.2 Experiments on the JEITA-HP database

In the following experiments, we always use 20 times of iterations ($p = 20$) to prepare training patterns for each epoch as well to measure the performance of our method. The training process is early stopped to avoid overfitting problem in case there is no improvement in the validation accuracy after 20 epochs. For evaluation, only a single $n$-tuple is sampled from each writer and fed into the trained network.

In the following sections, we present our experiments on the above databases. First, each local feature is used to build the writer identification model for evaluating its performance. Secondly, the two aggregation methods are employed on each local feature in order to find out the performance of different aggregation methods. Thirdly, the hyper-parameters, which consist of the number of writers, the number of character images for training and the number of characters for forming global features, are modified for evaluating the limitation of our networks and choosing the best hyper-parameters for later experiments. The above three stages are done on the JEITA-HP database.

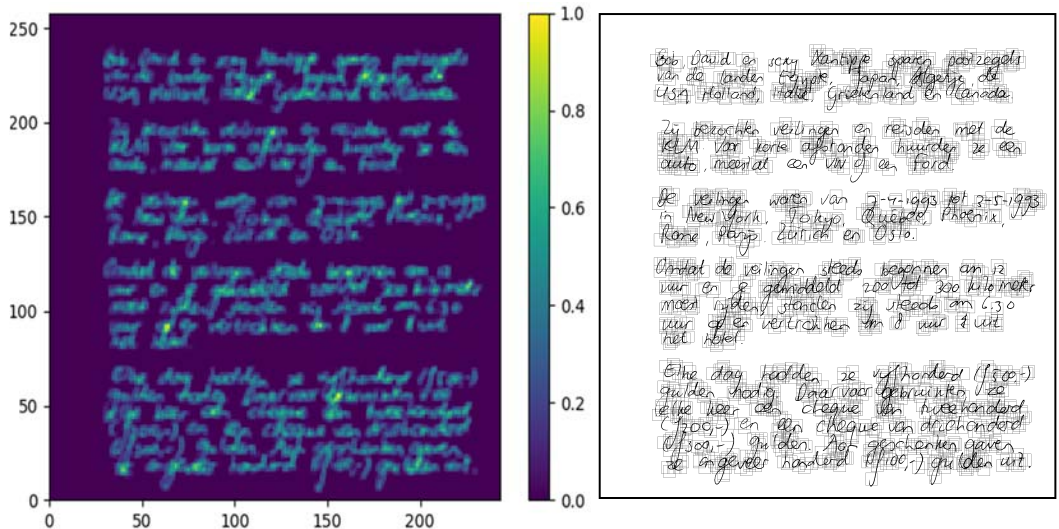

Figure 6: Sub-images extraction from a handwritten text page in the IAM database.

### 4.2.1 Experiments on Local features and Aggregation methods

In order to compare the local features and global features with different aggregation methods, we employed the experiments using 500 training characters of each writer to identify 100 writers, whose results are shown in Table 1. The first row of Table 1 shows the results of local features since there is no applied aggregation. Rows below the first row show the results with aggregations, thus global features. The global features are much better than the local features and the sub-region level based features are better than character level based features. The average aggregation method achieved the best accuracy of 99.97%. Thus, in the following experiments, we only apply the average aggregation method to form global features from local features.

The number of $K$ in KMA method is set by 10, 20, 40 and 50 for the sub-region level based features as well as the character level based features. The KMA method achieved higher accuracy rate than MA method for both the sub-region and character level based features since it preserves more information from local features than the MA method. The average aggregation could keep all the information from the local features so that its accuracy is always higher than other aggregation methods for all experiments.

### 4.2.2 Experiments on the tuple sizes

The following experiments determine the optimized value of $n$ for $n$-tuple. The size $n$ is set from 1 to 50 for both the sub-regions level based features (SF) and character level based features (CF) as shown in Fig. 7. In these experiments, we employ 500 characters for each writer in 100 writers and the average aggregation method since it is proved as the most efficient aggregation method. The sub-region level based features produce better recognition rate than the character level based features and achieve better identification rate even with a smaller tuple size of $n$. With only 10 character images input, the method identifies a writer with the accuracy of 99.09%. The highest identification rate is 99.97% by applying sub-region level based features with 20 character images.

Table 1: Accuracy (%) of local features and different aggregation methods on the JEITA-HP testing set.

| Aggregation \ Features | Sub-region level based | Character level based |
|---|---|---|
| None | 48.47 | 40.16 |
| AA | **99.97** | 99.78 |
| MA | 92.00 | 91.78 |
| KMA ($K$=10) | 99.53 | 98.56 |
| KMA ($K$=20) | 99.89 | 99.45 |
| KMA ($K$=40) | 99.82 | 99.78 |
| KMA ($K$=50) | 99.86 | 99.80 |

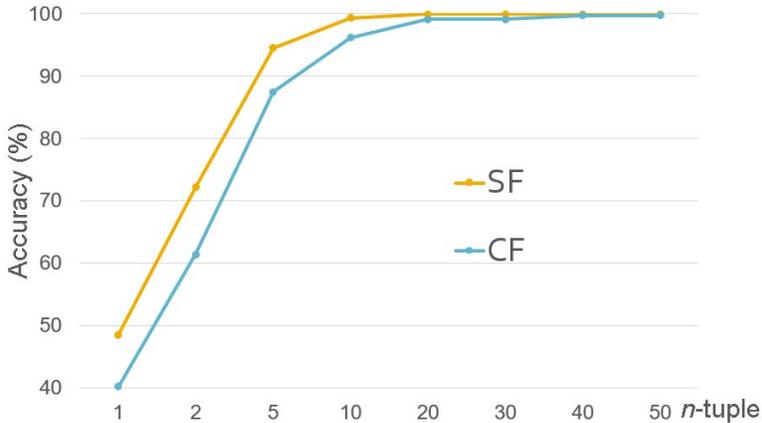

Figure 7: Accuracy (%) with different tuple sizes n on the JEITA-HP testing set.

### 4.2.3 Experiments on the number of training patterns per writer

To determine the smallest number of characters required for training, we did the experiments with the number of characters for training from 10 to 1000 of each writer to identify 100 writers. In these experiments, we use 20 for the tuple size *n* and the average aggregation method on sub-region level based features since these hyper-parameters are confirmed in the above sections. The results are presented in Fig. 8. Despite a small number of 50 characters for training, the performance of our method is 92.8% which implies that our method could extract text-independent writer-specific features using only a small number of training samples. The performance is improved by increasing the number of training samples. In the further experiments, we use 100 training samples for each writer since 100 characters are enough for achieving the accuracy of 97.52%.

### 4.2.4 Experiments on the number of writers

The final hyper-parameter is the number of writers. In these experiments, we change the number of writers from 2 to 400 and employ 20 for the tuple size *n*, 100 training characters and the average aggregation method on sub-region level based features. Fig. 8 shows the results of these experiments where the orange line presents the mean of accuracy in each experiment. The orange area limited by the upper and lower black lines in Fig. 8 expresses the range of variance in each experiment. The performance falls gradually while increasing the number of writers. However, our method achieves the accuracy of 93.82% using only 100 training characters for identifying 400 writers. Moreover, our method is evaluated 20 times for each experiment in order to obtain the variances. As shown in Fig. 9, the variances are approximate 1 point, which indicates that the training method could learn the writer-specific features independent from text.

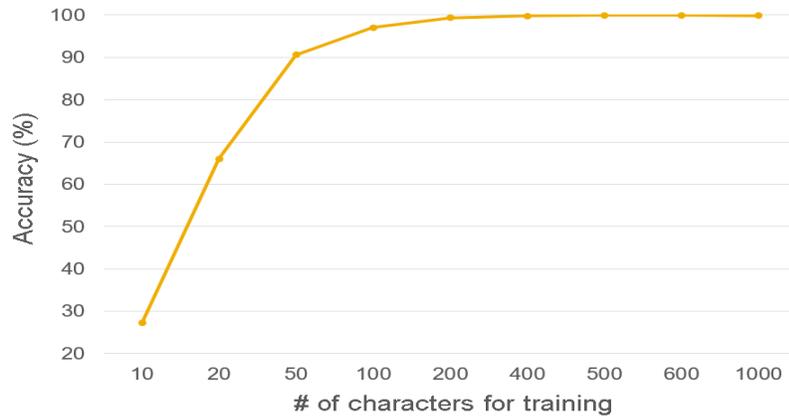

Figure 8: Accuracy (%) with different number of training characters on the JEITA-HP testing set.

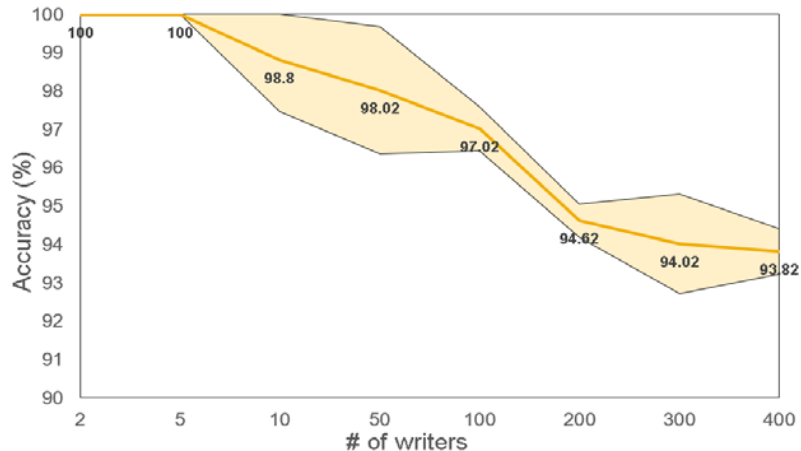

Figure 9: Accuracy (%) with different number of writers on the JEITA-HP testing set.

### 4.3 Experiments on the Firemaker and IAM databases

In the following sections, we present our experiments on the Firemaker and IAM databases with the best hyper-parameters from the above experiments. In addition, we compare the performance of our method with the results using the state-of-the-art handcrafted features and clustering methods [10]. Similarly as the experiments on the JEITA-HP database, we always use 20 times of iterations in each epoch ($p = 20$) for training and evaluation processes. The early stopped scheme is also applied as the above experiments.

#### 4.3.1 Performance on handwritten English text

For evaluation, only a single *n*-tuple is sampled from each writer and fed into the trained network. Our method, thus, does not require a large page or many characters for identifying writers during the evaluation process. Our method achieved the top-1 accuracy of 92.38% and top-10 accuracy of 97.67% for 250 writers of the Firemaker testing set. For 650 writers of the IAM testing set, our method reached 90.12% (top-1) and 97.82% (top-10). Although the tuple size *n* of 20 is used, the performance of our method is 91.5% for 900 writers of the Firemaker and IAM databases combination.

Since the samples of the Firemaker and IAM databases are cursively handwritten text, the extracted sub-images from them usually contain connected characters or only the sub-regions of connected characters, which requires the network to represent not only the features of isolated characters but also the features of connected characters. Thus, the performance of network on the Firemaker and IAM databases (92.38% for 250 writers) seems lower than on the JEITA-HP database (94.62% for 200 writers and 94.02% for 300 writers).

There might be also another reason for the less effect for these databases. The hyper-parameters were optimized for the handwritten Japanese text in JEITA-HP, so that they might not be optimal for English.

In addition, we employed five 20-tuples of sub-images for each test page to evaluate our method which gives better results than the top-1 accuracy by single 20-tuple (94.36% for five 20-tuples compared with 91.5% for single 20-tuple). It implies that we could extract and use more *n*-tuples of sub-images in order to achieve the better accuracy of writer identification in practice.

#### 4.3.2 Comparison with previously published results

Table 2 shows the comparisons between the method of Bulacu and Schomaker (2007) and our proposed method. For all the experiments on the English databases, the top-1 accuracy and top-10 accuracy of our proposed method are higher than the best single handcrafted feature (HF-single) and even the best combination of handcrafted features (HF-comb) in [10]. These results suggest that our proposed networks could represent writer-specific features for a large number of writers independent from text and even language.

Table 2: Writer identification accuracy (%) on the English databases.

| Database | | Firemaker | IAM | Firemaker + IAM |
|---|---|---|---|---|
| **Number of writers** | | 250 | 650 | 900 |
| **HF-single** | Top-1 | 81 | 81 | 80 |
| | Top-10 | 92 | 94 | 92 |
| **HF-comb** | Top-1 | 83 | 89 | 87 |
| | Top-10 | 95 | 97 | 96 |
| **Proposed network** | Top-1 | *92.38* | 90.12 | *91.50* |
| | Top-10 | 97.67 | 97.82 | 97.86 |
| | Top-1 by five 20-tuples | 93.56 | 93.14 | 94.36 |

## 5 Conclusions

This paper presented a CNN-based method for text-independent writer identification. Local features are extracted from a whole character pattern as well as its sub-regions and then they are aggregated to global features by three different methods. Random sampling was applied to create a large number of training patterns from a limited number of training samples for CNNs during the training process. The method learns writer-specific features automatically independent from text through end-to-end training and achieves 99.97% identification rate on the JEITA-HP database of Japanese handwritten character patterns for the task of identifying 100 writers. Our method achieved the accuracy higher than 92.8% which was trained by only 50 characters for 100 writers and 100 characters for 400 writers. This approach overcomes the difficulties of gathering handwritten character patterns of the same categories for writer identification.

For the experiments on the English databases (Firemaker and IAM), even though the number of writers is 900 and the imbalance of handwriting patterns of each writer in these databases, the accuracy results of our method are 91.5% higher than handcrafted features and clustering methods (80% when using a single handcrafted feature and 87% when combining their features). The high performance on the English databases with a large number of writers implies that the trained network represented the writer-specific features. It plays an important role in solving text-independent writer identification problem.

There remains work to pursue the limit and the scope of this method. Since we simply used the best hyper-parameters tuned for handwritten Japanese characters to the handwritten English text, the accuracy could be improved by tuning them for English. To evaluate our writer identification method for other languages or multiple languages [30] would also be interesting. In addition, it seems challenging to apply the method for identifying unknown writers or registering new writers.

## References

[1] R. Plamondon and G. Lorette, "Automatic signature verification and writer identification - the state of the art," *Pattern Recognit.*, vol. 22, no. 2, pp. 107–131, Jan. 1989, doi: 10.1016/0031-3203(89)90059-9.

[2] P. C. Chuang, "Machine verification of handwritten signature image," in *Proceedings of the International Conference On Crime Countermeasures Science and Engineering*, 1977, pp. 105–109.

[3] D. Bruyne and R. Forre, "Signature verification with elastic image matching," in *Proceedings of the International Carnahan Conference on Security Technology: Electronic Crime Countermeasures*, 1986, pp. 113–118.

[4] M. Yoshimura, F. Kimura, and I. Yoshimura, "Experimental Comparison of Two Types of Methods of Writer Identification," *IEICE Trans.*, vol. E65, no. 6, pp. 345–352, 1982.

[5] I. Yoshimura and M. Yoshimura, "Writer identification based on the arc pattern transformation," in *Proceedings of the 9th International Conference on Pattern Recognition*, 1988, pp. 35–37, doi: 10.1109/ICPR.1988.28166.

[6] S. N. Srihari, S. H. Cha, H. Arora, and S. Lee, "Individuality of handwriting: A validation study," in *Proceedings of the 6th International Conference on Document Analysis and Recognition*, 2001, pp. 106–109, doi: 10.1109/ICDAR.2001.953764.

[7] M. Sreeraj and S. M. Idicula, "A Survey on Writer Identification Schemes," *Int. J. Comput. Appl.*, vol. 26, no. 2, pp. 23–33, 2011, doi: 10.5120/3075-4205.

[8] H. E. S. Said, T. N. Tan, and K. D. Baker, "Personal identification based on handwriting," *Pattern Recognit.*, vol. 33, no. 1, pp. 149–160, 2000, doi: 10.1016/S0031-3203(99)00006-0.

[9] M. Bulacu, L. Schomaker, and L. Vuurpijl, "Writer identification using edge-based directional features," in *Proceedings of the International Conference on Document Analysis and Recognition*, 2003, pp. 937–941, doi: 10.1109/ICDAR.2003.1227797.

[10] M. Bulacu and L. Schomaker, "Text-independent writer identification and verification using textural and allographic features," *IEEE Trans. Pattern Anal. Mach. Intell.*, vol. 29, no. 4, pp. 701–717, 2007, doi: 10.1109/TPAMI.2007.1009.

[11] A. Bensefia, A. Nosary, T. Paquet, and L. Heutte, "Writer identification by writer's invariants," in *Proceedings of the 8th International Workshop on Frontiers in Handwriting Recognition*, 2002, pp. 274–279, doi: 10.1109/IWFHR.2002.1030922.

[12] C. Hertel and H. Bunke, "A Set of Novel Features for Writer Identification," *Audio- Video-Based Biometric Pers. Authentication*, vol. 2688, pp. 679–687, 2003, doi: 10.1007/3-540-44887-X_79.

[13] M. Bulacu and L. Schomaker, "A comparison of clustering methods for writer identification and

verification," in *Proceedings of the International Conference on Document Analysis and Recognition*, 2005, pp. 1275–1279, doi: 10.1109/ICDAR.2005.4.

[14] X. Wu, Y. Tang, and W. Bu, "Offline text-independent writer identification based on scale invariant feature transform," *IEEE Trans. Inf. Forensics Secur.*, vol. 9, no. 3, pp. 526–536, 2014, doi: 10.1109/TIFS.2014.2301274.

[15] V. Christlein, D. Bernecker, F. Hönig, and E. Angelopoulou, "Writer Identification and Verification Using GMM Supervectors," in *Proceedings of the IEEE Winter Conference on Applications of Computer Vision*, 2014, pp. 998–1005, doi: 10.1109/WACV.2014.6835995.

[16] D. Bertolini, L. S. Oliveira, E. Justino, and R. Sabourin, "Texture-based descriptors for writer identification and verification," *Expert Syst. Appl.*, vol. 40, no. 6, pp. 2069–2080, 2013, doi: 10.1016/j.eswa.2012.10.016.

[17] I. Siddiqi and N. Vincent, "Combining Contour Based Orientation and Curvature Features for Writer Recognition," in *Proceedings of the International Conference on Computer Analysis of Images and Patterns*, 2009, pp. 245–252, doi: 10.1007/978-3-642-03767-2_30.

[18] A. A. Brink, J. Smit, M. L. Bulacu, and L. R. B. Schomaker, "Writer identification using directional ink-trace width measurements," *Pattern Recognit.*, vol. 45, no. 1, pp. 162–171, 2012, doi: 10.1016/j.patcog.2011.07.005.

[19] U.-V. Marti, R. Messerli, and H. Bunke, "Writer identification using text line based features," in *Proceedings of the 6th International Conference on Document Analysis and Recognition*, 2001, pp. 101–105, doi: 10.1109/ICDAR.2001.953763.

[20] C. Djeddi, L.-S. Meslati, I. Siddiqi, A. Ennaji, H. El Abed, and A. Gattal, "Evaluation of Texture Features for Offline Arabic Writer Identification," in *Proceedings of the 11th IAPR International Workshop on Document Analysis Systems*, 2014, pp. 106–110, doi: 10.1109/DAS.2014.76.

[21] V. Christlein, D. Bernecker, A. Maier, and E. Angelopoulou, "Offline Writer Identification Using Convolutional Neural Network Activation Features," in *Proceedings of the German Conference on Pattern Recognition*, 2015, pp. 540–552, doi: 10.1007/978-3-319-24947-6_45.

[22] G. E. Hinton, S. Osindero, and Y.-W. Teh, "A Fast Learning Algorithm for Deep Belief Nets," *Neural Comput.*, vol. 18, no. 7, pp. 1527–1554, Jul. 2006, doi: 10.1162/neco.2006.18.7.1527.

[23] R. Collobert and J. Weston, "A unified architecture for natural language processing," in *Proceedings of the 25th international conference on Machine learning*, 2008, pp. 160–167, doi: 10.1145/1390156.1390177.

[24] K. He, X. Zhang, S. Ren, and J. Sun, "Deep Residual Learning for Image Recognition," in *Proceedings of the 29th IEEE Conference on Computer Vision and Pattern Recognition*, 2016, pp. 770–778, doi: 10.1109/CVPR.2016.90.

[25] S. Fiel and R. Sablatnig, "Writer Identification and Retrieval using a Convolutional Neural Network," in *Proceedings of the International Conference on Computer Analysis of Images and Patterns*, 2015, pp. 26–37, doi: 10.1007/978-3-319-23117-4_3.

[26] W. Yang, L. Jin, and M. Liu, "DeepWriterID: An End-to-End Online Text-Independent Writer Identification System," *IEEE Intell. Syst.*, vol. 31, no. 2, pp. 45–53, Mar. 2016, doi: 10.1109/MIS.2016.22.

[27] T. Kawatani and H. Shimizu, "Handwritten Kanji recognition with the LDA method," in *Proceedings of the 14th International Conference on Pattern Recognition*, 1998, vol. 2, pp. 1301–1305, doi: 10.1109/ICPR.1998.711940.

[28] L. R. B. Schomaker and L. Vuurpijl, "Forensic Writer Identification: A Benchmark Data Set and a Comparison of Two Systems," 2000.

[29] U. V. Marti and H. Bunke, "The IAM-database: An English sentence database for offline handwriting recognition," *Int. J. Doc. Anal. Recognit.*, vol. 5, no. 1, pp. 39–46, 2003, doi: 10.1007/s100320200071.

[30] D. Bertolini, L. S. Oliveira, and R. Sabourin, "Multi-script writer identification using dissimilarity," in *Proceedings of the International Conference on Pattern Recognition*, 2017, pp. 3025–3030, doi: 10.1109/ICPR.2016.7900098.